\documentclass[conference]{IEEEtran}
\IEEEoverridecommandlockouts
\usepackage{cite}
\usepackage{amsmath,amssymb,amsfonts}
\usepackage{algorithm}
\usepackage{algpseudocode}
\usepackage{graphicx}
\usepackage{subfigure}
\usepackage{textcomp}
\usepackage{graphicx}
\usepackage{placeins}
\usepackage{multirow}
\usepackage{tabularx}
\usepackage{float}
\usepackage{array}
\usepackage{caption} 
\usepackage{xcolor}

\def\BibTeX{{\rm B\kern-.05em{\sc i\kern-.025em b}\kern-.08em
    T\kern-.1667em\lower.7ex\hbox{E}\kern-.125emX}}
\begin{document}

\title{Real-Time Summarization of Twitter\\
}

\author{\IEEEauthorblockN{Yixin Jin*} 
\IEEEauthorblockA{
\textit{University of Michigan, Ann Arbor}\\
Ann Arbor, MI, 48109, USA \\
jinyixin@umich.edu}
\and
\IEEEauthorblockN{Meiqi Wang} 
\IEEEauthorblockA{
\textit{Brandeis University}\\
Waltham, MA, 02453, USA \\
meiqw@brandeis.edu}
\and
\IEEEauthorblockN{Meng Li} 
\IEEEauthorblockA{
\textit{Columbia University}\\
New York, NY, 10027, USA \\
ml4818@columbia.edu}
\and
\IEEEauthorblockN{Wenjing Zhou} 
\IEEEauthorblockA{
\textit{University of Michigan, Ann Arbor}\\
Ann Arbor, MI, 48109, USA \\
wenjzh@umich.edu}
\and
\IEEEauthorblockN{Yi Shen} 
\IEEEauthorblockA{
\textit{University of Michigan, Ann Arbor}\\
Ann Arbor, MI, 48109, USA \\
shenrsc@umich.edu}
\and
\IEEEauthorblockN{Hao Liu} 
\IEEEauthorblockA{
\textit{Northeastern University}\\
Shenyang, China \\
modiy.lu@gmail.com}
}

\maketitle

\begin{abstract}
In this paper, we describe our approaches to TREC
Real-Time Summarization of Twitter. We focus on real time push notification scenario, which requires a system monitors the stream of sampled tweets and returns the tweets relevant and novel to given interest profiles. Dirichlet score with and with very little smoothing (baseline) are employed to classify whether a tweet is relevant to a given interest profile. Using metrics including Mean Average Precision (MAP, cumulative gain (CG) and discount cumulative gain (DCG), the experiment indicates that our approach has a good performance. It is also desired to remove the redundant tweets from the pushing queue. Due to the precision limit, we only describe the algorithm in this paper.
\end{abstract}


\section{Introduction}
With the explosion of Internet and microblog, the information that one can access is enormous and is expected to grow faster and faster. In this circumstance, the need of information filtering is increasing \cite{zhu2021taming}. For example, a stock trader might want to know about the news or events about companies he has stock with and wish to receive notifications whenever something new happened for those companies. Other than the fact-checking studies\cite{Peng2024, read2023prediction, read2022prediction, lyu2022study, lyu2023attention, lyu2022multimodal, lyu2023backdoor} we might need in these cases, this information need is also considered in TREC Real-Time Summarization (RST) Track 

There are two different scenarios described in Real-Time Summarization (RST) Track \cite{xin2023self}. One is Push Notification: Given user's interest profile and sampled tweets stream, tweets that are identified as relevant and novel be pushed to the user in a timely manner.
The other one is Email Digest: Everyday, tweets with top relevance that day are digested in one email and pushed to the user.

We focus on the first scenario (push notification) in our project. To fulfill such kind of information need, our goal is to design a filtering system which is able to monitor the stream of tweets and find the tweets relevant to a given interest profile. On the other hand, it is desired to remove the redundant tweets. Some relevant tweets may contain the exact same information, and we want to keep only one tweet of those in our push queue according to the time order \cite{yang2024comparative}.   
Under the light of above discussion, our filtering system is designed to consist of two parts: classification of relevance and remove of redundancy. Given a interest profile, we will first convert it into a query. When a new tweet comes in, we will first calculate its similarity score with the query and classify it as relevant tweet or irrelevant tweet. Then if this new tweet is classified as a relevant tweet, we will compare it with relevant tweets recovered before, and classify it as novel tweet or redundant tweet. And we shall only keep novel tweets and push them to the users.
\section{Materials and Methods}
\subsection{Related Work}
\subsubsection{Smoothing in Language Model}
Tweets are usually short (140 words limit), as a result the language model \cite{xin2024vmt} estimated by maximum likelihood is limited. We need to assign positive probabilities for some words not observed. So we need smoothing. Zhai\cite{Zhai2001} suggests smoothing plays two roles in the ranking function: avoiding assigning zero probabilities
to words that have not occurred in a document and accommodating generation of common words in a query. The probabilities of unseen words are usually assigned by a reference model. 

Using Dirichlet Prior, the document language model $p(w|d)=\frac{|d|}{|d|+\mu}\frac{c(w,d)}{|d|}+\frac{\mu}{|d|+\mu}p(w|REF)$ where $|d|$ is the length of the document, $\mu$ is a smoothing parameter, $\frac{c(w,d)}{|d|}$ is the language model estimated by maximum likelihood and $p(w|REF)$ is the reference language model.

Another smoothing method is linear interpolation, or Jelinek-Mercer smoothing. $p(w|d)=(1-\lambda)\frac{c(w,d)}{|d|}+\lambda p(w|REF)$.

Comparing with the baseline with little smoothing, our results also show the importance of smoothing. 

\subsubsection{Clustering}
To remove the redundant tweets, clustering algorithms are needed. Clustering algorithms are unsupervised. Common clustering methods are k-means and agglomerative hierarchical clustering~\cite{chen2018data, bu2016attention}.

K-means minimizes the within class scatter
\begin{equation}
W(C)=\sum_{l=1}^k\sum_{i:C(i)=l}\|x_i-\bar{x_l}\|^2
\end{equation}
where k is the number of clusters, $C: \{1,2,\ldots,n\}\rightarrow\{1,2,\ldots,k\}$  is a function mapping data points to clusters, $x_i$ is the data point and $\bar{x_l}=\frac{1}{n_l}\sum_{j: C(j)=l}x_j$ is the center of the l-th cluster, where $n_l$ is the number of data points in the l-th cluster. This is a combinatorial optimization problem, so we optimize the function in an iterative way. We start with $k$ randomly selected data points and assume they are the centroids of the clusters. Then we assign every data point to a cluster whose
centroid is the closest to the data point. Next we recompute the centroid for each cluster. We repeat this process until the objective function converges. K-means is sensitive to initialization. In text clustering, we may not always get meaningful clusters.

In agglomerative hierarchical clustering, we group similar data points together in a bottom-up fashion and stop when some stopping criterion is met.

However, those algorithms need all the data in order to do clustering while our task is real-time \cite{chen2019claims}. Our redundancy removal method can handle tweets coming in a stream.

\subsection{Data and Preprocessing}
In order to evaluate our filtering system in the real time environment, we need the queries (interest profiles), a stream of tweets and the corresponding ground truth \cite{shen2024localization, li2024exploring, jiang2024disinformation}.

A collection of the interest profiles (queries) are provided by TREC 2016 \footnote{http://trecrts.github.io/TREC2016-RTS-topics.json} and we used the narrative part in the interest profile for the queries~\cite{li2024multi, li2023scigraphqa, yuan2024label, tan2024large}.

The ground truth can also be found on the TREC 2016 website \footnote{http://trec.nist.gov/data/rts/rts2016-qrels.txt}, in which relevance of sampled tweets (tweet id without tweet content) is given (0 for irrelevant, 1 for relevant, 2 for redundant). Furthermore, for each interest profiles, the relevant tweets are also clustered according to content \footnote{http://trec.nist.gov/data/rts/rts2016-batch-clusters.json}, and tweets in the same cluster are considered to be equivalent and contain the same information.

For tweets, Foreseer Group sampled tweets during the time span of the evaluation. The data from the Foreseer Group is really large (22G/day after compressing, 10\% sample of the original twitter stream) while the tweets for TREC is 1\% sample of the original twitter stream. We process the data to search for the intersection of the tweets with ground truth available and the tweets from Foreseer Group. Then we update the redundant labels in the ground truth as some relevant tweets may be missing. As a result, there are 27 queries in total, with about 1000 tweets per query on average.

\subsection{Methods}
In this section, we will discuss the methods we use to find the relevant tweets and remove the redundant tweets from all the relevant. The procedure we are going to present is summarized in Algorithm
\ref{alg:redundant}.

\subsubsection{Relevance}
\noindent As mentioned, given a query \(q\) and a tweet (document) \(d\), we will determine whether the tweet is relevant to query by calculating the similarity score with Dirichlet Smoothing \cite{chen2020optimal}, which is defined by
 \begin{equation}
 \small
 \label{eq:similarity_score}
 score(d,q)=\sum_{w \in d, w \in q}c(w,q)\log(1 + \frac{c(w,d)}{\mu \cdot p(w|C)}) + |q|\log \frac{\mu}{|d| + \mu}
 \end{equation}
 where $\mu$ is a parameter and the reference model \(p(w|C)\) is trained from the tweets on August 1 from Foreseer group. The resulting vocabulary size about $10^7$. The assumption we make here is that the language model of tweet from August 2nd to August 11th is similar to the language model on August 1. If the similarity score is above a threshold \(t\), then we will classify the tweet as relevant.   

In order to study the effect of smoothing, we will base our baseline method on the same similarity score but without smoothing, i.e. \(\mu = 0\). Because of computing difficulty (divide by 0 and log of 0), our practical baseline method is calculating the score with very little smoothing. 

\subsubsection{Redundancy}
\noindent With the consideration of novelty, we shall only push one tweet from a cluster, which means that the filtering system we build is expected to remove the redundant tweets among all the relevant \cite{xin2024mmap}. If a tweet is classified to be a relevant tweet, we can then calculate the cosine similarity between the new tweet and the previous novel tweets from each cluster. If the maximum cosine similarity is less than a threshold \(\theta\), then the tweet is regarded to be novel, and we shall keep this tweet. If the maximum cosine similarity is above the threshold \(\theta\), the tweet is considered to belong to the same cluster as an existing novel tweet and is redundant. For the consideration of efficiency, we don't keep redundant tweets. So for each relevant tweet, the number of cosine similarities we calculate is the number of current surviving tweets (which are considered to be relevant and novel).

However, in practice, we find the precision of our approach is already low (See Figure \ref{fig:bfig}, lower left plot). So we think that the evaluation of redundancy removal is not practical. The relevant tweets we retrieved form clusters and we keep only one tweet from each cluster, while the irrelevant tweets we retrieved are unlikely to be clustered. This will further reduce the precision and users will have a terrible experience.

\vfil 
\begin{algorithm}
\caption{Relevance and Redundancy Removal, \textbf{Input}: twitter stream, query $q$, threshold for relevance $t$, threshold for redundancy $\theta$  \textbf{Output}: relevant, novel tweets}
\begin{algorithmic}
\item[1] For a tweet $d$, calculate $score(d,q)$
\item[2] If $score(d,q)<t$, the tweet is irrelevant, drop $d$. Else, calculate if it's redundant 
\item[3] Calculate the cosine similarity $cos(d,d_i)$ where $d_i$ are one tweet from each current existing clusters
\item[4] If $\max_icos(d,d_i)<\theta$, $d$ is novel, we push and save $d$. Else, $d$ belong to a current existing cluster and is redundant. Drop $d$
\end{algorithmic}
\label{alg:redundant}
\end{algorithm}
\vfil 

\subsection{Experiment}
\subsubsection{Setting}
In order to simulate real time environment, for each query (interest profile), we organize the tweets (whose content and ground truth are both available) in the time order, and feed our filtering system one by one. In our method, the Dirichlet smoothing parameter is set to be $\mu=2500$ (see eq. \eqref{eq:similarity_score}). As mentioned, the practical baseline model we adapt has very little smoothing,  $\mu=0.000000001$. 

\subsubsection{Threshold \(t\)}
In this subsection, we discuss the effect of threshold \(t\) on precision and recall. The similarity score between the query \(q\) and tweet \(d\) is independent of \(t\), thus the method has different performance for different threshold \(t\). As we can observe from Figure \ref{fig:bfig}, for reasonable threshold, the precision is always low, which matches performance of other approaches for this problem \cite{Lin2016}. The reason for this phenomenon is that, among the huge stream of tweets, the fraction of relevant tweets is close to zero, actually for some query, there is no tweet is known to be relevant.  

For our method, when the threshold is in 3-5, the relationship between precision and recall is nearly linear. The precision and recall are not so sensitive to threshold either. So we set the threshold to be 4.5 here, as users may get annoying when there are too many irrelevant pushes.

For the baseline, when threshold is in the range of [-1000,-750], the precision is not improving much while the recall is degrading rapidly. The threshold is around the turning point of the precision-recall curve: (P=0.055, R=0.55), where the threshold is about -850.

\section{Results \& Discussion}
\subsection{Results}
In order to evaluate our method and compare it with the baseline method in detail, we calculate three common information retrieval metrics for both our method and the baseline. The first one is mean average precision (mAP)~\cite{gao2018solution,feng2022beyond,bu2021gaia,peng2023gaia}, which provides a single-figure measure of quality across recall levels. Among evaluation measures, mAP has been shown to have especially good discrimination and stability. For a single query (interest profile), mAP is the average of the precision values obtained for the set of top $k$ documents existing after each relevant document is retrieved, and this value is then averaged over all queries. The other two metrics we use are cumulative gain at 30 (CG@30) and discounted cumulative gain (DCG@30, log discount, base $r=2$). DCG is a measure of ranking quality often used to measure the effectiveness of web search engine algorithms or related applications~\cite{chen2022visual}. Using a graded relevance scale of documents in a search-engine result set, DCG measures the usefulness, or gain, of a document based on its position in the result list. The gain is accumulated from the top of the result list to the bottom, with the gain of each result discounted at lower ranks. 

\begin{figure}[!t]
\centering
\begin{subfigure}
\centering
\includegraphics[width = 0.23\textwidth]{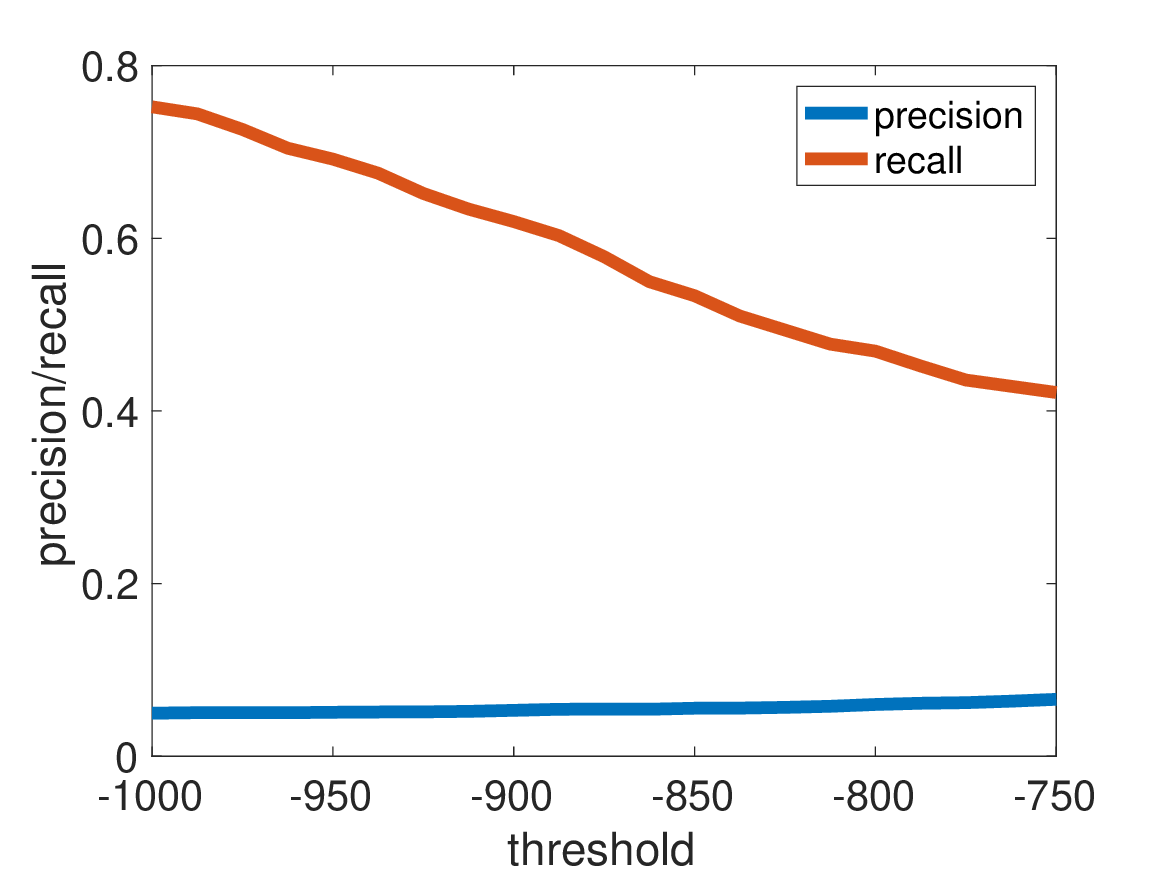}
\label{fig:bfig1}
\end{subfigure}
\begin{subfigure}
\centering
\includegraphics[width = 0.23\textwidth]{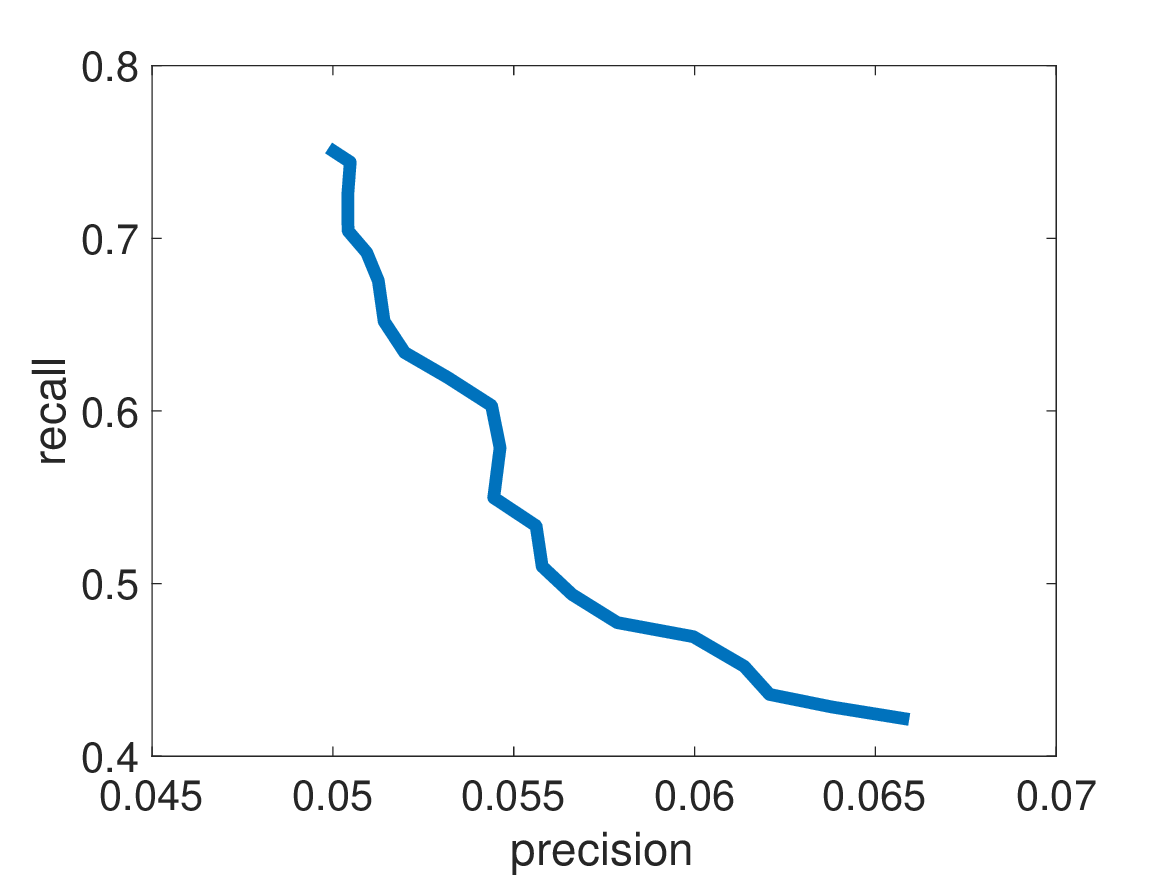}
\label{fig:bfig2}
\end{subfigure}
\begin{subfigure}
\centering
\includegraphics[width = 0.23\textwidth]{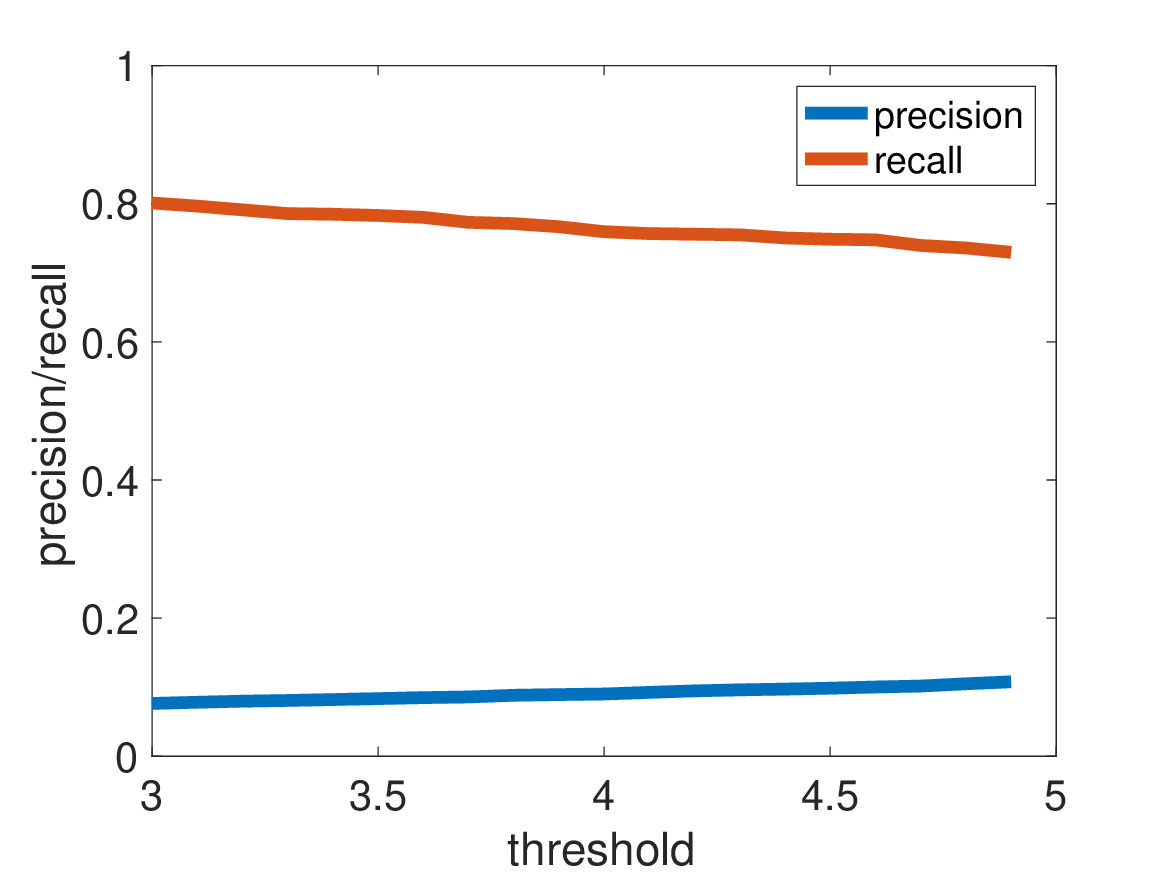}
\label{fig:bfig3}
\end{subfigure}
\begin{subfigure}
\centering
\includegraphics[width = 0.23\textwidth]{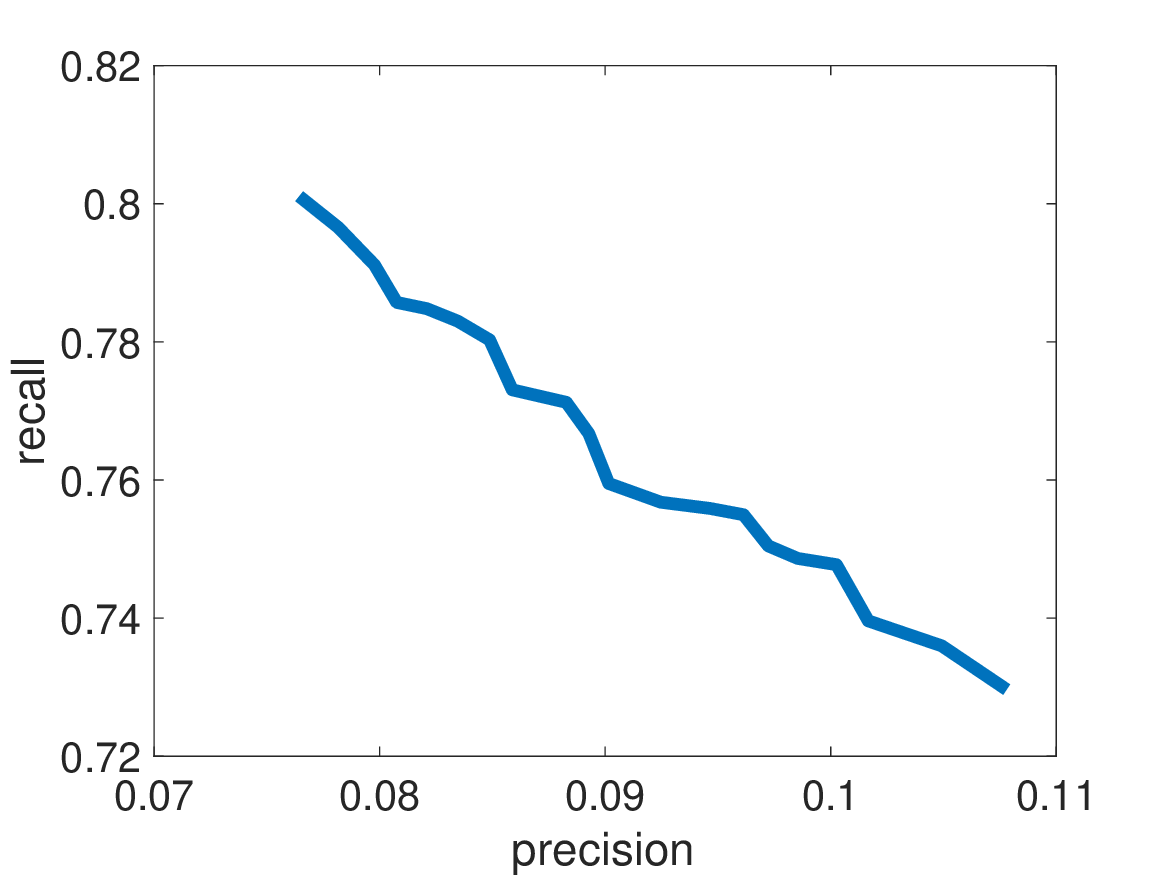}
\label{fig:bfig4}
\end{subfigure}
\caption{Precision and Recall of different thresholds for filtering system with and without smoothing. Top left panel: precision/recall versus threshold for baseline. Top right panel: recall versus precision for baseline. Bottom left panel: precision/recall versus threshold for filtering system with smoothing \(\mu = 2500\). Bottom right panel: recall versus precision for filtering system with smoothing \(\mu = 2500\).}
\label{fig:bfig}
\end{figure}

Table~\ref{table:formatting} reports the three metrics for our method and the baseline method. As can be seen from the table, our method outperforms the baseline method with respect to all three evaluation metrics by a large margin. The superiority of our methods can also be seen from the precision and recall curves (See Figure~\ref{fig:bfig}). It is clear that the precision-recall curve of our filtering system has a larger area under the curve than that of the baseline method. In addition, from the comparison of the threshold and precision/recall figures (See Figure~\ref{fig:bfig}, left panels), we can see that the values for the precision and recall of our filtering system are better than those of the baseline method. 

To provide a more comprehensive analysis of the performance metrics, we include 95\% confidence intervals for the reported improvements in mAP, CG@30, and DCG@30. We also performed paired t-tests to statistically validate the superiority of our method over the baseline, with p-values below 0.01 indicating statistically significant improvements.

\begin{table}
\centering
\caption{Results comparison between our method and the baseline, where the three metrics are calculated and averaged over all 27 queries.}
\label{table:formatting}
\begin{tabular}{|l|c|c|c|}
\hline
\textbf{Method} & \textbf{mAP} & \textbf{CG@30} & \textbf{DCG@30} \\
\hline
\hline
Baseline & 0.09 ± 0.02 & 0.88 ± 0.05 & 2.81 ± 0.20 \\
\hline
Our method & 0.17 ± 0.03 & 1.78 ± 0.07 & 5.42 ± 0.30 \\
\hline
\end{tabular}
\end{table}

\subsection{Discussion}
\subsubsection{Time Consumption}
\noindent When we calculate the similarity score in equation \eqref{eq:similarity_score}, we use vector representation (CountVectorizer in sklearn) of documents, and then the similarity score can be computed through an inner product form. As the vocabulary size is large ($10^7$), it takes about 1 second to calculate the similarity score for each incoming tweet. Considering the huge amount of tweets, this may not withstand the 1\% sample of the Twitter stream during the actual TREC competition.

The main reason behind this long time consumption is that, after smoothing, each query and tweet is converted to a long and dense vector (recall that our vocabulary size is about $10^7$). One modification is to truncate our vocabulary size. However, notice that the original queries and tweets vector are quite sparse. We hope that there are advanced computation approaches that can reduce the time consumption significantly.

Furthermore, while in this project we processed each query in series, in the actual TREC competition it might be better to process each query in a parallel manner.

\subsubsection{Future Work}
For future work, there are several directions we can pursue:
\begin{itemize}
\item \textbf{Email Digestion Scenario:} As we just need to digest all tweets into one email every day, this scenario can be categorized as a classical document ranking question. For the redundant removal, as we just need to cluster all tweets in a day that we regard as relevant, standard methods including K-means and agglomerative hierarchical clustering are expected to have good performance. This is because, in this scenario, redundancy removal does not need to be performed in real-time and we already have all the data at the end of the day.
\item \textbf{Precision Improvement:} Improve the precision of the retrieved documents such that it will be practical to remove the redundant tweets. The method might use other smoothing methods and similarity scores, such as KL divergence~\cite{Yao2016} and JS divergence~\cite{ChaoBei2016}. The threshold might be learned or updated every day~\cite{ChaoBei2016}. The use of emojis can also be adapted. Furthermore, this real-time summarization can be modeled as a reinforcement learning problem~\cite{Tan2017, liu2024adaptive, shen2024localization}. When the precision is high enough, we will implement the redundancy removal algorithm.
\item \textbf{Scalability:} Address the challenge of scaling up for higher tweet frequencies or larger user bases by exploring distributed computing frameworks such as Apache Spark and optimized data structures for fast retrieval. Implementing real-time processing pipelines and evaluating system performance under varying tweet volumes and different user interest profiles will also be crucial \cite{qu2024high}.
\end{itemize}
By addressing these aspects, we aim to further improve our method and make it more practical and scalable for real-world applications.

\section{Conclusion}
The research presented in this paper effectively demonstrates the superiority of a new method over a baseline approach in information retrieval tasks, showcasing significant improvements across all metrics such as mean average precision (mAP), cumulative gain at 30 (CG@30), and discounted cumulative gain at 30 (DCG@30). The detailed results, supported by precision-recall curves, confirm the new method’s enhanced ability to retrieve relevant documents more accurately. Conclusions drawn highlight the efficacy of the smoothing method used, which substantially boosts retrieval quality, and point to the need for efficiency improvements, given the high computational demands noted with large vocabulary sizes. This research holds substantial significance for both the academic community and practitioners by providing a proven approach that could enhance real-time data processing applications like social media analytics, improving both the relevance and precision of information retrieved. Additionally, it sets a clear path for future research, suggesting further exploration into optimization techniques and machine learning applications to refine retrieval processes.


\begin{thebibliography}{1}

\bibitem{ChaoBei2016}
Bei, C. and Hu, P. (2016) CCNU at TREC 2016 Real-Time Summarization Track. In: Proceedings of the 25th Text REtrieval Conference (TREC). Gaithersburg.

\bibitem{Lin2016}
Lin, J., Roegiest, A., Tan, L., McCreadie, R., Voorhees, E. M., \& Diaz, F. (2016) Overview of the TREC 2016 Real-Time Summarization Track. In: Proceedings of the 25th Text REtrieval Conference (TREC). Gaithersburg.

\bibitem{Tan2017}
Tan, H., Lu, Z. and Li, W. (2017) Neural network based reinforcement learning for real-time pushing on text stream. In: Proceedings of the 40th International ACM SIGIR Conference on Research and Development in Information Retrieval. Tokyo. pp. 913-916.

\bibitem{Yao2016}
Yao, L., Lv, C., Fan, F., Yang, J., \& Zhao, D. (2016) PKUICST at TREC 2016 Real-Time Summarization Track: Push Notifications and Email Digest. In: Proceedings of the 25th Text REtrieval Conference (TREC). Gaithersburg.

\bibitem{Peng2024}
Peng, X., Xu, Q., Feng, Z., Zhao, H., Tan, L., Zhou, Y., ... \& Zheng, Y. (2024) Automatic News Generation and Fact-Checking System Based on Language Processing. arXiv preprint arXiv:2405.10492.

\bibitem{Zhai2001}
Zhai, C., \& Lafferty, J. (2001) The dual role of smoothing in the language modeling approach. In: Proceedings of the Workshop on Language Models for Information Retrieval (LMIR). Pittsburgh. pp. 31-36.


\bibitem{zhu2021taming}Zhu, Z. \& Zhou, W. Taming heavy-tailed features by shrinkage. {\em International Conference On Artificial Intelligence And Statistics}. pp. 3268-3276 (2021)

\bibitem{xin2024vmt}Xin, Y., Du, J., Wang, Q., Lin, Z. \& Yan, K. VMT-Adapter: Parameter-Efficient Transfer Learning for Multi-Task Dense Scene Understanding. {\em Proceedings Of The AAAI Conference On Artificial Intelligence}. \textbf{38}, 16085-16093 (2024)

\bibitem{read2023prediction}Read, A., Zhou, W., Saini, S., Zhu, J. \& Waljee, A. Prediction of Gastrointestinal Tract Cancers Using Longitudinal Electronic Health Record Data. {\em Cancers}. \textbf{15}, 1399 (2023)

\bibitem{xin2024mmap}Xin, Y., Du, J., Wang, Q., Yan, K. \& Ding, S. MmAP: Multi-modal Alignment Prompt for Cross-domain Multi-task Learning. {\em Proceedings Of The AAAI Conference On Artificial Intelligence}. \textbf{38}, 16076-16084 (2024)

\bibitem{read2022prediction}Read, A., Zhou, W., Saini, S., Zhu, J. \& Waljee, A. PREDICTION OF GASTROINTESTINAL TRACT CANCERS USING LONGITUDINAL ELECTRONIC HEALTH RECORD DATA. {\em GASTROENTEROLOGY}. \textbf{162}, S1045-S1045 (2022)

\bibitem{liu2024adaptive}Liu, H., Shen, Y., Zhou, W., Zou, Y., Zhou, C. \& He, S. Adaptive speed planning for unmanned vehicle based on deep reinforcement learning. {\em ArXiv Preprint ArXiv:2404.17379}. (2024)

\bibitem{xin2023self}Xin, Y., Luo, S., Jin, P., Du, Y. \& Wang, C. Self-Training with Label-Feature-Consistency for Domain Adaptation. {\em International Conference On Database Systems For Advanced Applications}. pp. 84-99 (2023)

\bibitem{yang2024comparative}Yang, Q., Li, P., Shen, X., Ding, Z., Zhou, W., Nian, Y. \& Xu, X. A comparative study on enhancing prediction in social network advertisement through data augmentation. {\em ArXiv Preprint ArXiv:2404.13812}. (2024)

\bibitem{chen2019claims}Chen, S., Kong, N., Sun, X., Meng, H. \& Li, M. Claims data-driven modeling of hospital time-to-readmission risk with latent heterogeneity. {\em Health Care Management Science}. \textbf{22} pp. 156-179 (2019)

\bibitem{chen2020optimal}Chen, S., Lu, L., Zhang, Q. \& Li, M. Optimal binomial reliability demonstration tests design under acceptance decision uncertainty. {\em Quality Engineering}. \textbf{32}, 492-508 (2020)

\bibitem{shen2024localization}Shen, Y., Liu, H., Liu, X., Zhou, W., Zhou, C. \& Chen, Y. Localization through particle filter powered neural network estimated monocular camera poses. {\em ArXiv Preprint ArXiv:2404.17685}. (2024)

\bibitem{chen2018data}Chen, S., Lu, L., Xiang, Y., Lu, Q. \& Li, M. A data heterogeneity modeling and quantification approach for field pre-assessment of chloride-induced corrosion in aging infrastructures. {\em Reliability Engineering \& System Safety}. \textbf{171} pp. 123-135 (2018)

\bibitem{li2024exploring}Li, P., Yang, Q., Geng, X., Zhou, W., Ding, Z. \& Nian, Y. Exploring diverse methods in visual question answering. {\em ArXiv Preprint ArXiv:2404.13565}. (2024)


\bibitem{lyu2022study}
Lyu W, Zheng S, Ma T, et al. A Study of the Attention Abnormality in Trojaned BERTs[C]//Proceedings of the 2022 Conference of the North American Chapter of the Association for Computational Linguistics: Human Language Technologies. 2022: 4727-4741.

\bibitem{li2024multi}
Li S, Lin R, Pei S. Multi-modal preference alignment remedies regression of visual instruction tuning on language model[J]. arXiv preprint arXiv:2402.10884, 2024.

\bibitem{lyu2023attention}
Lyu W, Zheng S, Pang L, et al. Attention-Enhancing Backdoor Attacks Against BERT-based Models[C]//Findings of the Association for Computational Linguistics: EMNLP 2023. 2023: 10672-10690.

\bibitem{li2023scigraphqa}
Li S, Tajbakhsh N. Scigraphqa: A large-scale synthetic multi-turn question-answering dataset for scientific graphs[J]. arXiv preprint arXiv:2308.03349, 2023.

\bibitem{lyu2022multimodal}
Lyu W, Dong X, Wong R, et al. A multimodal transformer: Fusing clinical notes with structured EHR data for interpretable in-hospital mortality prediction[C]//AMIA Annual Symposium Proceedings. American Medical Informatics Association, 2022, 2022: 719.

\bibitem{lyu2023backdoor}
Lyu W, Zheng S, Ling H, et al. Backdoor Attacks Against Transformers with Attention Enhancement[C]//ICLR 2023 Workshop on Backdoor Attacks and Defenses in Machine Learning. 2023.

\bibitem{gao2018solution}
Gao Y, Bu X, Hu Y, et al. Solution for large-scale hierarchical object detection datasets with incomplete annotation and data imbalance[J]. arXiv preprint arXiv:1810.06208, 2018.

\bibitem{feng2022beyond}
Feng W, Bu X, Zhang C, et al. Beyond bounding box: Multimodal knowledge learning for object detection[J]. arXiv preprint arXiv:2205.04072, 2022.

\bibitem{chen2022visual}
Chen S, Lin C, Guan W, et al. Visual Encoding and Debiasing for CTR Prediction[J]. arXiv preprint arXiv:2205.04168, 2022.

\bibitem{bu2016attention}
Bu X, Pei M, Jia Y. Attention Estimation for Input Switch in Scalable Multi-display Environments[C]//Neural Information Processing: 23rd International Conference, ICONIP 2016, Kyoto, Japan, October 16–21, 2016, Proceedings, Part IV 23. Springer International Publishing, 2016: 329-336.

\bibitem{bu2021gaia}
Bu X, Peng J, Yan J, et al. Gaia: A transfer learning system of object detection that fits your needs[C]//Proceedings of the IEEE/CVF Conference on Computer Vision and Pattern Recognition. 2021: 274-283.

\bibitem{peng2023gaia}
Peng J, Chang Q, Yin H, et al. GAIA-Universe: Everything is Super-Netify[J]. IEEE Transactions on Pattern Analysis and Machine Intelligence, 2023, 45(10): 11856-11868.

\bibitem{jiang2024disinformation}
Jiang B, Tan Z, Nirmal A, et al. Disinformation detection: An evolving challenge in the age of llms[C]//Proceedings of the 2024 SIAM International Conference on Data Mining (SDM). Society for Industrial and Applied Mathematics, 2024: 427-435.

\bibitem{tan2024large}
Tan Z, Beigi A, Wang S, et al. Large Language Models for Data Annotation: A Survey[J]. arXiv preprint arXiv:2402.13446, 2024.

\bibitem{yuan2024label}
Yuan B, Chen Y, Tan Z, et al. Label Distribution Learning-Enhanced Dual-KNN for Text Classification[C]//Proceedings of the 2024 SIAM International Conference on Data Mining (SDM). Society for Industrial and Applied Mathematics, 2024: 400-408.

\bibitem{yuan2024label}
Yuan B, Chen Y, Tan Z, et al. Label Distribution Learning-Enhanced Dual-KNN for Text Classification[C]//Proceedings of the 2024 SIAM International Conference on Data Mining (SDM). Society for Industrial and Applied Mathematics, 2024: 400-408.

\bibitem{qu2024high}
Qu M. High Precision Measurement Technology of Geometric Parameters Based on Binocular Stereo Vision Application and Development Prospect of The System in Metrology and Detection. Journal of Computer Technology and Applied Mathematics, 2024, 1(3): 23-9.
\end{thebibliography}
\end{document}